\documentclass[11pt,a4paper]{article}
\usepackage{fontspec}
\usepackage{polyglossia}
\setdefaultlanguage{english}
\usepackage{amsmath,amssymb,amsfonts,mathtools}
\usepackage[margin=2.45cm]{geometry}

\title{\textbf{Bootstrap Theory of Representational Emergence (TBER)}\\[0.35em]
\large Explanatory Insufficiency, Transition Regimes, and the Emergence of New Representational Levels}
\author{Jacques Raynal$^{1,*}$ \and Pierre Slangen$^{2}$ \and Elsa Raynal$^{3}$ \and Jacques Margerit$^{4}$}
\date{}

\begin{document}
\maketitle

\begin{center}
\small
$^{1}$Laboratory of Bioengineering and Nanosciences (LBN), University of Montpellier, Montpellier, France\\
$^{2}$EuroMov Digital Health in Motion, University of Montpellier, IMT Mines Al\`es, Al\`es, France\\
$^{3}$Certified Sophrologist and Dental Assistant, Sensorimotor Practice, Montpellier, France\\
$^{4}$Emeritus Professor, University of Montpellier, Montpellier, France\\
$^{*}$Corresponding author: raynal.cab@gmail.com
\end{center}

\begin{abstract}
Representation learning has become a central paradigm in modern machine learning, enabling systems to move from handcrafted features to learned embeddings, latent spaces, foundation models, world models, and digital twins. Most research, however, focuses on how representations are optimized once a representational framework has been selected. Less attention has been given to the conditions under which a new level of representation becomes necessary. We introduce Version 4 of the Bootstrap Theory of Representational Emergence (TBER), a conceptual theory and research program describing how new representations arise when existing ones become explanatorily insufficient.

TBER identifies explanatory insufficiency as a positive epistemic signal for representational transition. A representation becomes insufficient not necessarily because it is false, but because the question posed to it exceeds what its current organization can make intelligible. The revised theory distinguishes two analytically separate dimensions. First, explanatory insufficiency may manifest descriptively, transformationally, or through failure of generalization. Second, the resulting transition may belong to different resolution regimes: a local-corrective regime, in which parameter or measurement refinement is sufficient; a representationally resolutive regime, in which re-representation substantially absorbs the identified insufficiency; or a structurally recurrent regime, in which extending or replacing the framework may remove a particular limitation while preserving the possibility of analogous limitations at the new level.

The bootstrap dynamic remains recursive: stabilized representations enable observation; observations reveal anomalies; persistent anomalies may expose explanatory insufficiency; insufficiency motivates candidate re-representations; discriminating tests constrain those candidates; and surviving representations enter provisional stabilization. Version 4 adds a regime-sensitive diagnosis between anomaly and transition, and clarifies that stabilization may be locally closing or recursively open. Formal mathematical cases such as Kaprekar's routine and G\"odelian incompleteness are used only as boundary examples of distinct transition regimes, not as proofs of TBER and not as models of biological or physical dynamics.

The framework concerns transitions between scientific, mathematical, or computational representations rather than transitions occurring within physical systems themselves. We discuss implications for representation learning, latent-space construction, foundation models, world models, adaptive biological systems, scientific discovery, and autonomous artificial intelligence. TBER suggests a design criterion for future intelligent systems: not only the ability to learn within a representation, but also the capacity to diagnose what kind of insufficiency has occurred, determine whether re-representation is warranted, generate discriminating tests, and recognize when a limitation is likely to be recurrent rather than eliminable.
\end{abstract}


\section{Introduction}
Representation learning is one of the central foundations of modern machine learning [14]. Its objective is not only to process observed data, but also to construct internal representations that make relevant structures, relations, and transformations more accessible to learning systems. This progression can be observed historically in the transition from handcrafted features to learned embeddings, latent spaces, foundation models, world models, and digital twins [13--17]. Most work in machine learning focuses on improving representations once a representational framework has already been selected.

Deep learning architectures, self-supervised learning, generative modeling, foundation models, and world-model architectures provide powerful methods for learning useful internal structures from large datasets [14--17]. These approaches have expanded the capacity of artificial systems to classify, predict, generate, simulate, transfer, and generalize. Yet a more fundamental question remains comparatively underdeveloped: \emph{when does a new level of representation become necessary?}

In many cases, the appearance of a new representation is justified retrospectively by improved performance, better compression, higher predictive accuracy, or increased transferability [13--15]. These criteria explain why a representation is useful after it has been introduced, but they do not by themselves explain why the prior representation became insufficient, why local correction was no longer enough, or whether the newly introduced representation should be expected to close the problem or merely move its boundary.

This problem is particularly visible in complex adaptive systems. Aggregated observable performance may fail to uniquely resolve experimental conditions, while richer multivariate representations may retain substantial overlap. Conversely, longitudinal transformations may remain inaccessible when observations are considered only through static representations [3,8,10,11,22,23]. A representation may therefore remain descriptively useful while becoming explanatorily insufficient. It may continue to summarize or organize observations without resolving their identity, transformation, generalization, or organizational relationships.

The same issue appears in contemporary artificial intelligence. Observable variables may be insufficient to capture relational structure; latent spaces may be insufficient to represent temporal transformation; predictive models may be insufficient to support simulation or planning; and world models may themselves become insufficient when systems must evaluate the adequacy of their own representational assumptions [14,16,17]. Each transition raises the same question: what makes the next representational level necessary?

TBER was introduced as a conceptual theory and research program for describing how new representations emerge when existing ones become explanatorily insufficient. Its central hypothesis is that representational innovation is driven not only by data availability, computational power, or scaling, but also by persistent explanatory gaps that cannot be resolved within the active representational framework. Explanatory insufficiency is therefore interpreted as a positive signal for representational transition.

Version 4 adds a further question that became necessary once the recursive character of TBER was taken seriously: \emph{are all explanatory insufficiencies of the same kind?} A limitation that disappears under a more adequate coordinate system is not epistemically equivalent to a limitation that reappears, in transformed form, whenever the framework is extended within a given class. The theory must therefore distinguish the \emph{manifestation} of an insufficiency from its \emph{resolution regime}.

We accordingly introduce two orthogonal classifications. The first concerns the form in which insufficiency becomes visible: descriptive, transformational, or generalization insufficiency. The second concerns the dynamics of response: local-corrective, representationally resolutive, or structurally recurrent. These classifications are not competing taxonomies. The first describes \emph{where} explanatory inadequacy appears; the second describes \emph{what happens when the representational system attempts to resolve it}.

The contribution remains theoretical rather than algorithmic. TBER does not introduce a new learning architecture, benchmark, empirical dataset, or optimization procedure. It proposes a meta-representational framework for interpreting when and why new representational levels become necessary, and how the character of an insufficiency may constrain the kind of transition that follows. The objective is not to propose a universal causal mechanism of scientific, mathematical, biological, or computational change, but to provide a shared interpretative structure whose validity and operational meaning must be examined separately within each domain.

The remainder of the article is organized as follows. Section 2 introduces the problem of representational accumulation. Section 3 defines explanatory insufficiency and introduces the distinction between manifestations and resolution regimes. Section 4 presents the revised bootstrap principle. Section 5 formalizes the five-stage transition model and its regime-sensitive refinement. Section 6 situates TBER relative to epistemology, complex systems, representation learning, world models, and formal mathematical boundary cases. Section 7 discusses implications for machine learning and autonomous artificial intelligence. Section 8 examines adaptive biological systems. Section 9 presents the founding empirical trajectory. Section 10 discusses originality, scope, and limitations. Section 11 outlines future directions and concludes.

\section{The Problem of Representational Accumulation}
Contemporary science and artificial intelligence are characterized by a rapid multiplication of representational levels. In machine learning, this evolution can be seen in the progression from manually engineered features to learned embeddings, latent spaces, foundation models, world models, and digital twins [14--17]. In biomedical and digital health research, traditional clinical variables are increasingly complemented by biological markers, behavioral measurements, digital biomarkers, multimodal sensor data, and predictive models [18--21]. This diversification has greatly expanded the capacity of scientific and artificial systems to describe, classify, predict, simulate, and intervene.

Each new representational level may reveal aspects of a system that were inaccessible at previous levels. Latent spaces may reveal hidden relational structures; foundation models may support transfer across domains; world models may represent possible transformations; and digital twins may integrate multiple scales of observation into dynamic simulations. However, multiplication of representations does not automatically produce greater intelligibility. A model may improve predictive accuracy without explaining the organization that generated the observed data. A latent space may separate observations without clarifying the relations that make the separation meaningful. A digital twin may reproduce system behavior without making explicit why a particular level of representation was required.

This distinction between representational performance and explanatory necessity is central to TBER. In many scientific domains, new representations are justified retrospectively because they improve classification, prediction, compression, simulation, or decision-making. This pragmatic criterion is legitimate but incomplete. It explains why a representation is useful once available, but not why the prior representation became insufficient or whether the new one resolves the relevant limitation.

Representations may therefore accumulate without an explicit account of the conditions governing their emergence. Observable variables are supplemented by latent variables; static descriptions by dynamic models; predictive systems by world models; clinical measurements by digital biomarkers. Yet the logic of transition between these levels often remains implicit. TBER distinguishes representational multiplication from the scientific necessity of introducing a new representational level.

The problem is not the existence of multiple representations. Complex systems may legitimately require several complementary levels. The problem arises when representational multiplication is interpreted as progress in itself. More variables, larger datasets, deeper models, or more abstract latent spaces do not necessarily guarantee better explanation. They may increase descriptive power while leaving the relevant organization insufficiently understood.

Version 4 adds that representational accumulation can fail in two distinct ways. First, it can fail because the active framework is merely under-refined: a better estimate, improved calibration, additional observations, or parameter adjustment may be sufficient. Second, it can fail because the scientific question requires a representational relation not expressible or not accessible within the current framework. In the first case, a transition of representation may be unnecessary. In the second, local refinement cannot by itself restore explanatory adequacy.

This distinction prevents TBER from treating every anomaly as evidence of emergence. The theory is concerned with the threshold at which representational refinement ceases to be a sufficient response and re-representation becomes scientifically warranted.

\section{Explanatory Insufficiency as an Epistemic Event}
\subsection{Definition}
Explanatory insufficiency describes a situation in which a representation remains useful for some descriptive, predictive, or operational purposes but no longer provides a satisfactory account of observations, relations, transformations, or organizational properties relevant to the question under investigation.

In ordinary scientific practice, model insufficiency is often interpreted as a temporary weakness. The model may be corrected, extended, calibrated, or replaced. Such adjustments are legitimate and often necessary. However, they do not exhaust the meaning of insufficiency. In some cases, a persistent limitation indicates not merely that a model requires local correction, but that the current representational level has reached the boundary of its explanatory domain.

TBER adopts this second interpretation only after plausible local explanations have been considered. A representation is not necessarily false because it becomes insufficient. Classical mechanics remains useful within its domain despite being inadequate for relativistic phenomena. Observable performance metrics may remain useful for comparison even when they are insufficient for understanding the organizations that produce those performances. A latent space may remain useful for classification while becoming insufficient for representing temporal trajectories or counterfactual transformations.

We therefore define explanatory insufficiency as follows:

\begin{quote}
\textbf{Explanatory insufficiency} occurs when a representation remains capable of organizing, describing, predicting, or otherwise supporting inquiry within part of its domain, but fails persistently to make intelligible a relation, distinction, transformation, or organizational property that has become relevant to the scientific or computational question, and when this failure cannot be adequately attributed to ordinary noise, missing data, parameter error, or correctable implementation defects.
\end{quote}

This definition is broader than prediction error. A model may predict accurately while remaining explanatorily poor, and may predict imperfectly while revealing meaningful organizational structure. TBER therefore distinguishes predictive performance from explanatory adequacy.

\subsection{Manifestations of explanatory insufficiency}
Version 3 proposed three useful forms of explanatory insufficiency. Version 4 retains them, but explicitly interprets them as \emph{manifestations} rather than resolution regimes.

\textbf{Descriptive insufficiency} occurs when the active representation fails to preserve distinctions relevant to the question. Different organizations may map to similar observable states, or scalar aggregation may remove relations needed for discrimination.

\textbf{Transformational insufficiency} occurs when a representation can describe states but fails to represent changes, trajectories, temporal dependencies, counterfactual transformations, or transformations between representational states.

\textbf{Generalization insufficiency} occurs when a representation remains effective within its initial domain but fails under transfer, distribution shift, new tasks, or contextual demands that require relations not adequately encoded in the active representation.

These manifestations answer the question: \emph{where does the insufficiency become visible?}

\subsection{Resolution regimes}
Version 4 introduces a second axis: the regime under which an insufficiency can be addressed. This axis answers a different question: \emph{what kind of transition is required, and what happens after the transition?}

\textbf{Local-corrective regime.} The observed inadequacy can be removed without changing the representational level. Parameter adjustment, improved measurement, additional data, calibration, error correction, or ordinary model refinement is sufficient. In this regime, TBER predicts no representational emergence.

\textbf{Representationally resolutive regime.} The inadequacy cannot be adequately removed within the current representation, but a re-representation introduces relations or coordinates that substantially absorb the identified insufficiency. The new representation may therefore close the specific explanatory gap that motivated the transition, while remaining provisional with respect to future questions.

\textbf{Structurally recurrent regime.} A transition can eliminate a particular limitation, but the relevant class of representations retains a principled capacity to generate analogous limitations at the new level. The transition increases explanatory or formal reach without guaranteeing closure. The new representation is therefore stabilized locally while the bootstrap remains structurally open.

These regimes should not be read as an ordinal scale from ``weak'' to ``strong.'' They characterize different structures of response. A descriptive insufficiency, for example, may be local-corrective in one case and representationally resolutive in another. A transformational insufficiency may be resolutive or recurrent depending on the domain and the constraints on admissible representations.

\subsection{Why the distinction matters}
Without this distinction, the claim that ``insufficiency motivates a new representation'' is too coarse. A scientifically useful theory of representational transition must distinguish at least three questions:
\begin{enumerate}
\item Is the anomaly genuine or attributable to ordinary error?
\item If genuine, does it indicate a limitation of the current representation or only of its current parametrization?
\item If re-representation is required, should the transition be expected to close the identified gap or merely to move the boundary of adequacy?
\end{enumerate}

The third question is the principal addition of Version 4. It makes the recursive character of TBER more than a generic statement that ``new problems may appear later.'' It requires the theory to consider whether recurrence is contingent on future discoveries or is linked to the structure of the representational class itself.

\subsection{Explanatory insufficiency as a positive signal}
Explanatory insufficiency is not a property of the observed phenomenon alone. It characterizes the adequacy of a representational framework relative to a question. This gives insufficiency a methodological role: it provides a criterion for deciding when a new representational level may be justified. Without such a criterion, new representations risk being introduced merely because they are technically available or fashionable.

What resists explanation is therefore not automatically noise or failure. It may indicate relational, temporal, organizational, counterfactual, or formal structure that exceeds the current representational framework. Peirce emphasized surprising facts in abductive reasoning [1]; Bachelard described scientific progress through epistemological obstacles [2]; Canguilhem showed how exceptional conditions can reveal hidden dimensions of living systems [3,4]; Simondon described individuation as a resolution of tensions [7]. TBER shifts the focus from the emergence of hypotheses alone to the emergence of representational levels.

The question is not only: ``Which hypothesis explains this observation?'' It is also: ``Which representational level is required for the observation, relation, or transformation to become intelligible, and what kind of limitation is being encountered?''

\section{The Revised Bootstrap Principle}
TBER proposes that representational evolution follows a recursive bootstrap process in which each representational level creates the conditions for the emergence of subsequent levels. The term \emph{bootstrap} is used in an epistemic sense: a representational system progressively transforms its own conditions of intelligibility.

Existing representations enable observation; observation reveals limitations; limitations may motivate new representations; and new representations redefine what can subsequently be observed, discriminated, or explained. The process operates recursively rather than linearly. Representational emergence is not simple accumulation of information. Additional data, variables, or computational resources do not automatically produce a new representational level.

The revised principle is:

\begin{quote}
\textbf{Every representation creates the conditions under which its own limits may become detectable; when a persistent limitation cannot be adequately repaired within the active representational level, it becomes a candidate signal for re-representation. The resulting transition must then be evaluated not only for explanatory gain, but also for its resolution regime: local closure, representational resolution, or structural recurrence.}
\end{quote}

This principle concerns the evolution of representational frameworks. It does not imply that observed physical or biological systems themselves undergo equivalent representational transitions. The distinction is essential. TBER is a theory of representational adequacy and transition, not a universal ontology of the systems being represented.

Representations are not only tools for understanding phenomena; they also determine which insufficiencies can become visible. A representation allows certain questions to be asked while rendering others inaccessible or poorly posed. Consequently, each representational framework simultaneously expands knowledge and generates new zones of ignorance.

Knowledge therefore does not progress merely by adding facts to an existing framework. Periods of representational stability may be punctuated by transitions in which the framework itself changes. TBER also differs from purely revolutionary accounts of scientific change because it does not require abrupt replacement. Multiple representational levels may coexist and remain useful within different explanatory domains.

Version 4 sharpens the meaning of provisionality. A representation may be provisional in at least two senses. It may be \emph{contingently provisional}: later evidence may reveal an unforeseen limitation. Or it may be \emph{structurally provisional}: the representational class is known, in the relevant sense, not to guarantee complete closure. TBER does not assume that every domain exhibits the second form. The distinction is introduced precisely to prevent the recursive nature of the theory from being interpreted as a universal claim of necessary endlessness.

For artificial intelligence, this refinement has a direct implication. A representationally intelligent system should not only detect that its current representation has failed. It should attempt to infer whether the failure is correctable within the current level, requires re-representation, or belongs to a recurrent regime in which ``solving'' one instance should not be confused with eliminating the class of limitations.

\section{A Five-Stage Model of Representational Transition}
The bootstrap principle is operationalized through five stages. Version 4 preserves the five-stage macroscopic architecture but adds regime diagnosis and closure assessment inside the transitions between stages.

\subsection{Stage 1: Stabilized Observation}
The process begins with a stabilized representational framework. Observations are organized according to an accepted set of concepts, variables, models, or representations. Stabilization does not imply completeness. It indicates that the representation is sufficiently effective within its current explanatory domain.

A stabilized representation may support measurement, prediction, classification, simulation, or reasoning. Its stability is practical and epistemic rather than absolute. At this stage, the system has no sufficient evidence that the active representation must be replaced.

\subsection{Stage 2: Anomaly Detection}
As observations accumulate or questions change, certain phenomena may become difficult to reconcile with the existing framework. Anomalies may take the form of unexpected observations, unexplained variability, contradictory results, instability under changing conditions, systematic prediction errors, unresolved overlap, or persistent residual structure.

An anomaly does not immediately invalidate a representation. Many anomalies can be accommodated through local correction. Version 4 therefore emphasizes that anomaly detection is not yet a representational diagnosis.

\subsection{Stage 3: Recognition and Typing of Explanatory Insufficiency}
The third stage is the critical transition point. Anomalies cease to be interpreted as isolated irregularities and begin to be recognized as indicators of a deeper limitation. The problem is no longer merely whether an observation is unusual, but whether the representational framework can make it intelligible.

The transition from anomaly to explanatory insufficiency requires evidence that the anomaly is persistent, structured, relevant to the scientific question, and resistant to reasonable correction within the current framework. Isolated errors, measurement noise, insufficient data, poor parameter estimation, or software defects should not be treated as evidence for representational emergence.

Version 4 adds two diagnostic tasks at this stage:
\begin{enumerate}
\item identify the manifestation of insufficiency (descriptive, transformational, generalization, or another domain-specific form);
\item formulate a hypothesis about its resolution regime (local-corrective, representationally resolutive, or structurally recurrent).
\end{enumerate}

The second task is initially provisional. The regime may become identifiable only after candidate re-representations have been tested.

\subsection{Stage 4: Representational Emergence}
Once explanatory insufficiency has been recognized, a new representational framework may emerge. The new representation introduces concepts, variables, relations, structures, coordinates, or organizational principles unavailable or inaccessible in the previous framework.

The essential feature of emergence is not novelty alone but the capacity to reduce an identified insufficiency. However, emergence must not be conflated with validation. A new representation remains a candidate until its explanatory adequacy has been constrained by theoretical or empirical discrimination.

\subsection{From Emergence to Stabilization: Re-representation, Discriminating Tests, and Selection}
Three operations may occur between Stage 4 and Stage 5.

\textbf{Problem re-representation.} The insufficiency identified at Stage 3 may require not only a new representation of the observations but also reformulation of the problem. Variables, relations, transformations, or constraints that were secondary in the previous framework may become central.

\textbf{Discriminating test.} A candidate representation becomes scientifically informative when it generates consequences that distinguish it from competing representations or from explanations available within the previous framework. The relevant test is not merely confirmatory; it reduces representational ambiguity.

\textbf{Representational selection.} The outcome of discriminating tests constrains the space of admissible representations. Candidates that fail to account for relevant observations lose explanatory adequacy; candidates that survive remain viable for stabilization.

The transition can be summarized as
\begin{equation}
R_t \rightarrow A_t \rightarrow I_t^{(m,r)} \rightarrow P_{t+1} \rightarrow T_{t+1} \rightarrow \Sigma_{t+1} \rightarrow R_{t+1},
\end{equation}
where $R_t$ denotes the currently stabilized representation, $A_t$ an anomaly, $I_t^{(m,r)}$ a recognized explanatory insufficiency characterized provisionally by manifestation $m$ and resolution regime $r$, $P_{t+1}$ a candidate re-representation, $T_{t+1}$ a discriminating test, $\Sigma_{t+1}$ representational selection, and $R_{t+1}$ a provisionally stabilized representation.

This notation does not constitute a mathematical model of TBER. It is a bookkeeping formalism that makes explicit the information that a future operational implementation would need to track.

\subsection{Stage 5: Provisional Stabilization and Closure Status}
Following representational emergence, a candidate framework may stabilize only after sufficient theoretical or empirical constraint has reduced relevant ambiguity. Observations are reorganized according to the new framework and explanatory coherence is restored sufficiently for investigation or application.

Version 4 introduces the notion of \emph{closure status}. A stabilized representation can have one of three provisional statuses with respect to the insufficiency that triggered the transition:

\textbf{Local closure:} the apparent insufficiency was resolved without a genuine level transition; the active representation was corrected or refined.

\textbf{Representational closure:} the new representation resolves the identified insufficiency within the scope of the current question, although future unrelated insufficiencies may occur.

\textbf{Recursive openness:} the new representation resolves or bypasses the particular insufficiency while preserving a principled or empirically persistent possibility of analogous insufficiencies at the new level.

No claim is made that these statuses can always be determined conclusively. Their purpose is to prevent stabilization from being treated as a homogeneous endpoint.

\subsection{Recursive Dynamics}
The five stages constitute a recursive dynamic that can occur at multiple scales and across multiple domains. A scientific discipline may experience several representational transitions over decades. A computational system may successively construct new internal representations as it encounters increasingly complex tasks. An analytical pipeline may move from scalar summaries to multivariate embeddings and then to longitudinal transformations.

The revised cycle can be summarized as follows: stabilized representations generate observations; observations reveal anomalies; anomalies may reveal explanatory insufficiencies; insufficiencies are typed; candidate representations emerge; discriminating tests constrain them; surviving representations stabilize; and their closure status determines whether the relevant explanatory problem is locally closed or remains recursively open.

The significance of the model lies in its generality, but that generality is also a constraint. TBER does not imply that all domains share identical mechanisms. It proposes a common structure for asking how representational change becomes necessary.

\section{Relation to Existing Epistemological, Computational, and Formal Frameworks}
\subsection{Scientific Discovery and Epistemological Rupture}
The role of explanatory insufficiency in TBER shares affinities with classical theories of scientific discovery. Peirce emphasized surprising observations in abductive reasoning [1]. Bachelard argued that scientific development proceeds through overcoming epistemological obstacles rather than simple accumulation [2]. Foucault emphasized historically contingent conditions under which knowledge becomes possible [5,6].

TBER does not replace these accounts. It focuses specifically on the emergence of representational levels. Its question is not only how hypotheses change, but under what conditions the representational framework in which hypotheses are formulated becomes inadequate.

\subsection{Biological Adaptation and Organizational Change}
Varela's theory of biological autonomy described living systems as self-producing organizations maintained through recursive processes [8]. Kauffman emphasized self-organization in biological complexity [10]. These frameworks focus on organizational change in biological systems. TBER concerns a different but related question: when do such changes expose the limits of the scientific or computational frameworks used to describe them?

The distinction is important. A biological reorganization is not itself a representational transition in the TBER sense. The representational transition occurs in the analytical, scientific, or computational framework attempting to make the biological process intelligible.

\subsection{Complex Systems and Self-Organization}
Kelso's dynamic patterns and Newell's constraints-based approach emphasize emergent coordination [11,12]. TBER complements these approaches by examining how representational frameworks change when existing descriptions cannot adequately account for the organization or transformation being studied.

\subsection{Representation Learning and Machine Learning}
Representation learning concerns how systems discover useful internal structures for prediction, classification, and generalization [14]. TBER addresses a prior or higher-order question: under what conditions does a new representational level become necessary?

This distinction separates learning \emph{within} a representation from diagnosing the inadequacy of the representation itself. Version 4 extends this further by requiring diagnosis of whether inadequacy is parametric, representationally resolutive, or structurally recurrent.

\subsection{Foundation Models and World Models}
Foundation models support broad transfer through large-scale representational structures [15]. World models attempt to represent dynamics and possible transformations rather than merely encode observations [16]. TBER interprets such developments as potential representational transitions motivated by limitations of earlier frameworks, without claiming that historical development actually followed TBER explicitly.

World models are particularly important because they illustrate that increased representational scope does not eliminate the problem of adequacy. A model capable of simulation may still fail to recognize when its assumptions have become insufficient. TBER therefore concerns not only richer representation, but meta-representational evaluation.

\subsection{Autonomous Machine Intelligence}
LeCun has argued that future machine intelligence will require world models, predictive capabilities, planning, and reasoning [17]. TBER extends this trajectory by proposing that autonomous systems may require mechanisms for detecting when their own representational frameworks become insufficient.

Version 4 adds that such a system should not automatically respond to every insufficiency by expanding the model. It should distinguish correction from re-representation and recognize when a transition may be recursively open.

\subsection{Formal Mathematical Boundary Cases}
Formal mathematical examples provide unusually controlled environments for clarifying the distinction between resolution regimes. They are useful precisely because the relevant objects and transformations can be specified without experimental noise. However, they should be treated as boundary cases for conceptual discrimination, not as empirical validations of TBER.

Kaprekar's four-digit routine provides an example of a representationally resolutive transition [26]. The routine is completely describable at the level of individual four-digit states, but its structure becomes more transparent after ordering the digits $a\ge b\ge c\ge d$ and observing that one iteration depends on the differences
\begin{equation}
x=a-d, \qquad y=b-c,
\end{equation}
through
\begin{equation}
K=999x+90y.
\end{equation}
The original representation is not false; it is redundant relative to the transformation. Re-representation exposes a reduced state structure and renders the fixed-point dynamics more intelligible. The relevant insufficiency can therefore be substantially absorbed by a better representation.

G\"odel's incompleteness results provide a fundamentally different boundary case [27,28]. For sufficiently expressive, effectively axiomatized formal theories satisfying the relevant consistency conditions, there are statements that cannot be decided within the theory, and the theory cannot in general establish its own consistency by its own resources in the manner excluded by the second incompleteness theorem. Extending the theory may decide a particular previously undecidable statement, but an appropriately extended theory that remains within the scope of incompleteness can acquire new undecidable statements. This is not a failure of coordinate choice analogous to the Kaprekar case. It illustrates a structurally recurrent regime in which increasing formal reach does not provide generic closure.

TBER does not derive either mathematical result, and neither result proves TBER. Their value is discriminatory: they show that the phrase ``representation becomes insufficient'' can correspond to transition structures that are mathematically non-equivalent. One can be resolved by a more adequate representation of the same transformation; the other demonstrates a class-level limitation on complete formal closure.

This comparison motivates the Version 4 distinction between representational resolution and structural recurrence.

\subsection{A Distinct Contribution}
The originality of TBER lies not in claiming that anomalies, model revision, conceptual change, or representation learning are new phenomena. It lies in making transitions between representational levels themselves the object of analysis and proposing explanatory insufficiency as a criterion for their necessity.

Version 4 adds a second claim: the nature of an insufficiency constrains the regime of transition. Representational emergence should therefore be studied not only in terms of whether a new representation appears, but also in terms of whether the triggering insufficiency was locally correctable, representationally resolutive, or structurally recurrent.

\section{Implications for Representation Learning, Foundation Models, and Autonomous Artificial Intelligence}
\subsection{Representation Learning as a Sequence of Transitions}
Representation learning is typically described as the process through which a system discovers internal structures facilitating prediction, classification, generation, or decision-making [14]. From the perspective of TBER, representational development can also be interpreted as a sequence of responses to insufficiencies encountered at previous levels.

Handcrafted features may become inadequate for relational complexity; learned embeddings may fail to represent temporal transformation; static prediction may become insufficient for planning; and a world model may become insufficient when its own assumptions must be revised. The point is not that each historical development was caused by explanatory insufficiency, but that insufficiency provides a criterion for deciding when such a transition is scientifically warranted.

\subsection{Latent Spaces as Intermediate Representational Levels}
Latent spaces reorganize observations according to relations not directly visible in the original variables. They can reveal similarity, structure, and transformation more coherently than raw observables. Yet latent spaces possess finite domains of adequacy. They may organize states while failing to model temporal evolution, causal structure, counterfactual reasoning, or long-horizon planning.

Version 4 adds that failure of a latent space must first be diagnosed. If the problem is poor estimation, local correction may suffice. If the geometry itself excludes relevant relations, re-representation may be required. If every extension within the chosen model class reproduces a particular limitation, the system may be facing a recurrent regime.

\subsection{Foundation Models and Representational Generalization}
Foundation models extend representation beyond task-specific learning by constructing large-scale structures capable of transfer across multiple domains [15]. They may be interpreted as responses to the generalization insufficiency of narrowly specialized representations. Nevertheless, broad transfer does not eliminate representational insufficiency. Limitations may remain in reasoning, planning, causal inference, interpretability, or autonomous adaptation.

The important implication of Version 4 is that scaling cannot be assumed to change the resolution regime. A larger model may repair a local insufficiency, expose a representational one, or leave a recurrent limitation untouched. TBER therefore separates \emph{capacity increase} from \emph{representational transition}.

\subsection{World Models and Internal Simulation}
World models attempt to model dynamics generating observations [16]. Their objective is not only to represent states but also possible transformations between states. Within TBER, world models illustrate a transition from descriptive to transformational representation.

Yet a world model may simulate transformations without evaluating the adequacy of its own representational assumptions. Version 4 therefore treats world-model adequacy as provisional and introduces the possibility that some limitations are recursively open rather than solvable by model enlargement alone.

\subsection{Toward Regime-Aware Autonomous Representational Systems}
Most contemporary AI systems learn within predefined architectural and representational constraints. They rarely determine autonomously when a new representational level is required. A TBER-inspired autonomous system would require at least the following capacities:
\begin{enumerate}
\item detect persistent explanatory anomalies;
\item distinguish ordinary error from representational limitation;
\item identify the manifestation of insufficiency;
\item estimate whether the limitation is local-corrective, representationally resolutive, or potentially recurrent;
\item construct candidate re-representations;
\item reformulate the problem to generate discriminating observations or tests;
\item select among candidates under empirical, logical, predictive, or domain-specific constraints;
\item assign a provisional closure status after stabilization.
\end{enumerate}

This list defines a research program rather than an existing computational architecture.

\subsection{Autonomous Scientific Discovery}
Scientific progress often occurs when observations expose limitations that cannot be resolved through parameter adjustment alone. Future systems for autonomous scientific discovery may therefore require mechanisms for identifying explanatory insufficiency and exploring alternative representations rather than searching only within a fixed representational space.

Version 4 makes one additional requirement explicit: autonomous discovery should not equate successful re-representation with final explanatory closure. A scientifically competent system should preserve uncertainty about whether the new framework is merely locally adequate or whether its class of representations is subject to recurrent limitations.

\subsection{Representational Intelligence}
These considerations motivate the concept of \emph{representational intelligence}. Traditional machine learning focuses primarily on learning within representations. Representational intelligence additionally involves the capacity to evaluate, transform, test, select, and replace representations when necessary.

Version 4 refines the concept as follows:
\begin{quote}
\textbf{Representational intelligence} is the capacity of a system to reason about the adequacy of the representational structures through which observations and problems are organized, to distinguish correctable from representationally significant insufficiencies, to construct and discriminate among alternative representations, and to estimate whether resulting stabilization is locally closing or recursively open.
\end{quote}

This form of intelligence would operate at a higher organizational level than ordinary parameter learning while remaining dependent on empirical and computational constraints.

\section{Adaptive Biological Systems as a Domain of Representational Emergence}
Adaptive biological systems provide a particularly informative domain for examining TBER because they continuously face changing environmental conditions, internal perturbations, developmental transformations, and functional constraints. Their organization often cannot be understood through static descriptions alone.

\subsection{The Limits of Observable Performance}
Observable performance does not necessarily reflect internal organization. Similar outcomes may emerge from distinct coordination strategies, and substantial organizational transformations may occur while aggregate performance remains stable [3,4,11,12]. A representation based exclusively on observable outcomes may therefore remain descriptively accurate while becoming insufficient for understanding the organization producing those outcomes.

Version 4 adds a methodological caution: such non-identifiability does not by itself establish a representationally resolutive or structurally recurrent regime. Biological variability, measurement limitations, and underdetermination may all contribute. The resolution regime must be inferred through discriminating analyses rather than assumed from overlap alone.

\subsection{Adaptation as Organizational Reconfiguration}
Living systems may reorganize interactions among components, redistribute functional roles, modify coordination patterns, or exploit unused degrees of freedom [8,10]. From the perspective of TBER, such reorganizations may reveal that the current scientific representation has become insufficient. The biological transformation and the representational transition remain analytically distinct.

A previously sufficient description may become unable to account for adaptive behavior, while the new representation embeds rather than invalidates the old one. TBER therefore favors representational stratification over simple substitution.

\subsection{Longitudinal Displacement as an Emergent Analytical Level}
Recent investigations of adaptive locomotor systems provide an illustration [22--25]. Initial analyses focused on aggregated performance metrics. Their relative rankings were sensitive to score construction, normalization, weighting, and treatment of unavailable values. A richer multivariate representation was introduced to preserve relationships removed by scalar aggregation.

The resulting embedding retained substantial overlap among observational probes and did not reveal independently separated condition-specific clusters. The additional representation therefore preserved multivariate information without resolving static condition identity. This persistent non-identifiability motivated a change in analytical question: from static identification to longitudinal representational displacement.

M1--M2 observations were projected into a common PCA coordinate system and condition-level centroid displacement was compared [23]. The displacement was interpreted only as a retrospective and representation-dependent descriptor of longitudinal change, not as a direct measurement of physiological viability, biological stability, or therapeutic suitability.

From a Version 4 perspective, this sequence illustrates primarily descriptive and transformational insufficiencies. It does not establish structural recurrence in the G\"odelian sense, nor does it show that the biological domain is intrinsically non-closing. The empirical evidence supports successive analytical reformulations, not a theorem about the impossibility of a final biological representation.

\subsection{Biological Systems as Natural Laboratories of Representational Transition}
Evolution, development, learning, adaptation, compensation, and rehabilitation generate situations in which existing representations may become insufficient. Biological systems therefore provide rich environments for studying representational transitions. But TBER must maintain a strict distinction between changes in the biological system and changes in the framework used to represent that system.

\subsection{Toward a General Theory of Adaptive Representation}
The relevance of adaptive biological systems extends beyond biology. Artificial systems increasingly display learning, planning, self-organization, and environmental interaction. As their complexity increases, they may encounter representational limitations analogous in form, though not necessarily in mechanism, to those observed in biological research.

Version 4 proposes that cross-domain comparison should focus on transition structure rather than superficial similarity. The useful question is not whether two systems ``adapt'' in the same way, but whether the representational frameworks used to understand them encounter comparable manifestations and resolution regimes of insufficiency.

\section{The Founding Empirical Trajectory: From Static Non-Identifiability to Internal Approximation}
TBER did not originate from a purely philosophical reflection. It emerged from a sequence of empirical investigations conducted on adaptive locomotor systems, where successive explanatory insufficiencies motivated changes in analytical representation and scientific question [22--25].

The initial objective was to compare experimental gait conditions using aggregated biomechanical performance metrics. Observable variables derived from gait measurements were combined into composite scores intended to summarize apparent performance. Reconstructed analysis subsequently showed that relative ranking depended on score construction, including variable orientation, normalization, weighting, and treatment of unavailable values [22].

Observable performance therefore remained informative, but it did not provide a unique or weighting-independent identification of the experimental conditions. This first insufficiency motivated the introduction of a richer multivariate representation. The purpose was not to infer a hidden physiological organization, but to determine whether relationships removed by scalar aggregation became more informative in a reduced-dimensional embedding.

The revised multivariate analysis showed substantial overlap among the observational probes and did not reveal independently separated condition-specific clusters [22]. The richer representation retained information but did not resolve static non-identifiability. This unresolved ambiguity generated a new question: even when conditions overlap at a given session, do their condition-level representations exhibit different changes between measurement sessions?

This second insufficiency motivated the transition from static representation to longitudinal analysis. M1 and M2 observations were projected into a common PCA coordinate system and condition-level centroid displacement was compared [23]. The resulting displacement was interpreted only as a retrospective, representation-dependent proxy for longitudinal reorganization.

The longitudinal analysis itself remained retrospective. It described M1--M2 displacement after both sessions had been acquired. It did not determine whether the observed coordinate transformation could be represented computationally. This third insufficiency motivated a transition toward internal approximation. A simplified supervised model was used to test whether the observed M1--M2 transformation in the selected PCA representation could be approximated within the same single-subject dataset [24].

The objective was deliberately restricted. The model did not predict an unseen patient, recover physiological dynamics, or establish clinical viability. It tested only whether an already observed representation-dependent transformation contained enough regularity to be approximated internally. The succession of these analytical reformulations motivated the development of the bootstrap framework itself [25].

The trajectory can be summarized as
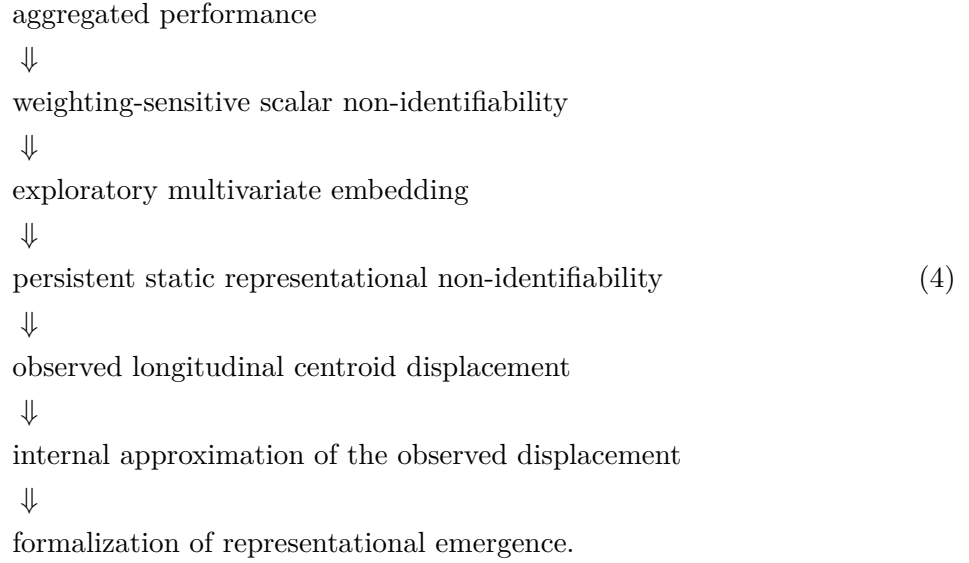
\begin{equation}
\begin{aligned}
&\text{aggregated performance} \\
&\Downarrow \\
&\text{weighting-sensitive scalar non-identifiability} \\
&\Downarrow \\
&\text{exploratory multivariate embedding} \\
&\Downarrow \\
&\text{persistent static representational non-identifiability} \\
&\Downarrow \\
&\text{observed longitudinal centroid displacement} \\
&\Downarrow \\
&\text{internal approximation of the observed displacement} \\
&\Downarrow \\
&\text{formalization of representational emergence.}
\end{aligned}
\end{equation}

Version 4 clarifies the interpretation of this trajectory. Each transition demonstrates that a previous analytical level became insufficient for a newly relevant question. The sequence does not establish that each insufficiency was structurally recurrent, nor that successive levels progressively unveil an intrinsic hidden physiological reality. Each level is an analytical construction that reorganizes what can be asked and observed.

The empirical trajectory therefore supports the core TBER claim that unresolved or negative findings can be scientifically productive. It also illustrates why Version 4 requires regime diagnosis: repeated transitions in one empirical program are not, by themselves, evidence that recurrence is a necessary property of the domain.

\section{Discussion: Originality, Scope, and Limitations of TBER}
\subsection{From Representation Learning to Representational Emergence}
A first contribution of TBER is the distinction between representation learning and representational emergence. Representation learning explains how increasingly effective representations can be acquired. TBER asks why a new representational level becomes necessary. It therefore concerns transitions between representational frameworks rather than optimization within a fixed framework.

\subsection{Explanatory Insufficiency as a Positive Signal}
A second contribution is the interpretation of explanatory insufficiency as a positive epistemic signal. Prediction errors, anomalies, and inconsistencies are often treated as failures to be minimized. TBER proposes that persistent insufficiency may reveal the boundary of a representational domain and thereby indicate where new representational possibilities should be sought.

Version 4 adds that this signal must be typed. A positive signal for inquiry is not automatically a signal for representational replacement. Local-corrective insufficiencies should be corrected locally. Representational emergence becomes warranted only when the inadequacy concerns the active level itself.

\subsection{The Nature of Insufficiency Constrains the Transition Regime}
The principal theoretical addition of Version 4 is the proposition that not all explanatory insufficiencies have the same transition structure. Some are correctable without re-representation. Some are substantially resolved when the problem is expressed in a more adequate representational space. Others may recur across extensions of a representational class.

This leads to the revised TBER proposition:
\begin{quote}
\textbf{The necessity of representational change is determined not only by the existence of explanatory insufficiency, but also by the type of insufficiency and by the resolution regime under which that insufficiency can be addressed.}
\end{quote}

The proposition is conceptual, not yet a theorem or operational decision rule.

\subsection{A General Framework Across Domains}
TBER does not depend on a specific scientific discipline, computational architecture, or biological mechanism. Similar transition structures may be identified in scientific discovery, adaptive biological systems, machine learning, formal mathematics, and complex-systems research. This generality is both a strength and a challenge.

TBER does not claim that these transitions are generated by identical mechanisms. The distinction between Kaprekar and G\"odel in Section 6 is itself evidence against such flattening. Their value lies in clarifying transition regimes, not in suggesting that arithmetic routines, formal theories, biological systems, and machine-learning models are ontologically equivalent.

\subsection{Relationship to Existing Theories}
TBER should not be interpreted as a replacement for theories of learning, adaptation, self-organization, formal logic, or scientific change. Representation learning explains how representations are acquired. Self-organization explains how collective structures emerge. Viability theory explains how systems maintain trajectories under constraints. Formal logic establishes results about derivability and consistency. Scientific epistemology examines how knowledge evolves.

TBER addresses a complementary question: how and why representational levels change when existing ones become insufficient, and what kind of transition follows from the character of that insufficiency.

\subsection{Emergence, Discrimination, and Provisional Stabilization}
Representational emergence and representational validation are distinct operations. Explanatory insufficiency can motivate a new candidate representation without guaranteeing that the candidate is adequate. Re-representing the problem may alter which observations or tests become relevant. Provisional stabilization is therefore the outcome of constrained selection, not the immediate consequence of novelty.

Version 4 strengthens this distinction by adding closure assessment. A representation can survive discrimination and still remain recursively open with respect to the class of limitation that triggered its emergence.

\subsection{Limitations}
Several limitations must be acknowledged.

First, TBER remains a conceptual theory rather than a formal mathematical model. It does not provide an implementable pipeline for detecting, generating, discriminating, selecting, or typing representations automatically.

Second, the distinction among resolution regimes is currently conceptual. No general quantitative criterion is provided for deciding whether a limitation is locally correctable, representationally resolutive, or structurally recurrent. In many empirical domains, recurrence may be impossible to establish definitively.

Third, formal mathematical boundary cases clarify conceptual distinctions but do not validate the theory across empirical domains. G\"odelian incompleteness cannot be generalized to biological or machine-learning representations without additional formal justification. Likewise, a successful change of coordinates in a finite arithmetic routine does not establish that empirical representational transitions are generally resolutive.

Fourth, the empirical trajectory motivating TBER originates from a specific domain involving adaptive locomotor systems [22--25]. Broader empirical validation remains necessary.

Fifth, the distinction between a correctable anomaly and a genuine representational limitation may remain ambiguous. Future work must develop criteria capable of distinguishing these situations more rigorously.

Sixth, real systems may exhibit overlapping transitions, partial reorganizations, hybrid representations, and nested explanatory levels that do not fit neatly into a single five-stage sequence.

Finally, TBER does not claim that all representational evolution is open-ended by necessity. Version 4 explicitly distinguishes contingent provisionality from structural recurrence to avoid turning recursion into an unfalsifiable universal principle.

\subsection{Toward a Science of Representational Transitions}
Despite these limitations, TBER suggests that representational emergence itself may constitute a legitimate object of scientific investigation. Scientific inquiry traditionally focuses on observations, hypotheses, models, mechanisms, and predictions. TBER proposes that transitions between representational levels deserve comparable attention.

Version 4 sharpens the research program. A science of representational transitions should study not only when a representation becomes insufficient, but also how insufficiency manifests, which corrective actions are attempted, whether re-representation is required, how candidates are discriminated, and whether stabilization closes or recursively reopens the relevant explanatory problem.

\section{Future Directions and Conclusion}
\subsection{Operationalizing Explanatory Insufficiency}
Future work should investigate measurable indicators of explanatory insufficiency, including persistent prediction failure, instability under distribution shift, loss of transferability, unresolved overlap, sensitivity to preprocessing or weighting, failure to represent temporal transformations, and structured residual patterns.

The immediate objective is to distinguish ordinary model error from representational limitation.

\subsection{Operationalizing Manifestation and Resolution Regime}
Version 4 adds a second operational objective: determine whether an identified insufficiency is descriptive, transformational, generalization-related, or another domain-specific form, and then estimate its likely resolution regime.

Potential indicators of a local-corrective regime include disappearance of the anomaly under calibration, improved sampling, parameter adjustment, or correction of measurement error without changing the representational variables or relations.

Potential indicators of a representationally resolutive regime include the introduction of new variables or coordinates that reduce structured ambiguity, expose stable invariants, enable previously inaccessible transformations, or produce robust discriminating tests unavailable in the old framework.

Potential indicators of structural recurrence would require stronger evidence. These might include formal impossibility results, repeated reappearance of the same class of limitation across systematically expanded representations, or theoretically justified boundary conditions showing that the limitation belongs to the representation class rather than a particular instance. Empirical repetition alone should not be considered sufficient.

\subsection{Operationalizing Discriminating Tests and Representational Selection}
Future work should define criteria for determining whether a newly generated representation produces genuinely discriminating consequences rather than merely redescribing the same observations. Candidate indicators include creation of new observables, improved separation among previously indistinguishable hypotheses, reduction of structured residual ambiguity, robustness under independent tests, and transfer across contexts.

A computational implementation of TBER would therefore require not only a detector of explanatory insufficiency, but also a mechanism for proposing alternative problem representations, generating tests that maximize discrimination among those alternatives, and deciding when evidence is sufficient for provisional stabilization.

\subsection{Toward Autonomous Representational Evolution}
Current machine-learning systems learn powerful representations within predefined architectural constraints. They rarely determine for themselves when a new representational level is required. TBER suggests that future systems may need explicit mechanisms for detecting explanatory insufficiency and generating alternative representational frameworks.

Version 4 further suggests that autonomous representational evolution requires epistemic restraint. A system should not interpret every successful transition as final closure. It should preserve a model of its own closure status and distinguish uncertainty about future insufficiency from evidence of structural recurrence.

\subsection{Applications to Scientific Discovery}
Future work could apply TBER retrospectively to historical cases in physics, mathematics, biology, neuroscience, and medicine in order to determine whether the five-stage model and the regime distinction capture recurring patterns of representational change.

Such analyses should be comparative and discriminating. The aim is not to redescribe every scientific change as a TBER cycle, but to identify cases where the theory makes nontrivial distinctions between correction, re-representation, and recurrence.

\subsection{Applications to Adaptive Biological Systems}
Further studies in locomotion, posture, development, neuroplasticity, pathology, and rehabilitation could test whether transitions from observable performance to multivariate representation, longitudinal analysis, and internal approximation follow the bootstrap sequence. Such work must distinguish analytical representations, physiological states, longitudinal descriptors, and clinically validated outcomes.

A central Version 4 question is whether empirical transitions are genuinely representationally resolutive or merely improve a local description while leaving the relevant ambiguity open.

\subsection{Conclusion}
The Bootstrap Theory of Representational Emergence proposes that new levels of representation become scientifically necessary when existing frameworks remain useful but become explanatorily insufficient for observations, relations, transformations, or organizational properties relevant to a question.

Version 4 preserves the recursive five-stage architecture while introducing a major refinement: explanatory insufficiency has both a \emph{manifestation} and a \emph{resolution regime}. Descriptive, transformational, and generalization insufficiencies describe where inadequacy appears. Local-corrective, representationally resolutive, and structurally recurrent regimes describe what kind of response is required and whether the resulting stabilization is locally closing or recursively open.

The revised bootstrap sequence is therefore not merely
\begin{equation}
\text{insufficiency} \rightarrow \text{new representation},
\end{equation}
but rather
\begin{equation}
\begin{aligned}
\text{anomaly} &\rightarrow \text{diagnosis} \rightarrow \text{typed insufficiency} \rightarrow \text{candidate re-representation} \\
&\rightarrow \text{discrimination} \rightarrow \text{selection} \rightarrow \text{closure assessment}.
\end{aligned}
\end{equation}

Formal mathematical boundary cases clarify why this distinction is necessary. Some representational limitations can be substantially absorbed by an appropriate re-representation; others may be recurrent properties of a representational class. These cases do not prove TBER, but they constrain its vocabulary and prevent all transitions from being treated as structurally equivalent.

The principal implication for future artificial intelligence is correspondingly stronger. Intelligent systems may benefit not only from the ability to learn representations, but from the ability to diagnose the limits of those representations, determine what kind of insufficiency has occurred, construct and test alternatives, and recognize whether a successful transition closes the relevant explanatory problem or only moves its boundary.

TBER should therefore be regarded as a theory of representational transitions rather than a theory of the observed systems themselves. Its central research question is no longer only \emph{when must representation change?} Version 4 adds a second: \emph{what kind of limitation made the change necessary, and what kind of closure can the new representation legitimately claim?}

\section*{Author Contributions}
Jacques Raynal conceived the Bootstrap Theory of Representational Emergence (TBER), developed the theoretical framework, synthesized the empirical trajectory underlying the theory, and wrote the original manuscript. Pierre Slangen contributed to the conceptual development of the framework, participated in the interpretation of the representational transitions described in the manuscript, and critically revised the text. Elsa Raynal contributed to the conceptual discussion of adaptive systems and representational emergence and revised the manuscript. Jacques Margerit contributed to the scientific supervision of the project, theoretical validation of the framework, and critical revision of the manuscript. All authors approved the manuscript.

\section*{Funding}
No external funding was received for this work.

\section*{Conflicts of Interest}
The authors declare no conflict of interest.

\section*{References}
\small
1. Peirce CS. How to Make Our Ideas Clear. \textit{Popular Science Monthly}. 1878;12:286--302.

2. Bachelard G. \textit{La formation de l'esprit scientifique}. Paris: Vrin; 1938.

3. Canguilhem G. \textit{Le normal et le pathologique}. Paris: Presses Universitaires de France; 1966.

4. Canguilhem G. \textit{The Normal and the Pathological}. New York: Zone Books; 1991.

5. Foucault M. \textit{Les mots et les choses}. Paris: Gallimard; 1966.

6. Foucault M. \textit{L'arch\'eologie du savoir}. Paris: Gallimard; 1969.

7. Simondon G. \textit{L'individuation \`a la lumi\`ere des notions de forme et d'information}. Grenoble: J\'er\^ome Millon; 2005.

8. Varela FJ. \textit{Principles of Biological Autonomy}. New York: North Holland; 1979.

9. Aubin JP. \textit{Viability Theory}. Boston: Birkh\"auser; 1991.

10. Kauffman SA. \textit{The Origins of Order: Self-Organization and Selection in Evolution}. Oxford: Oxford University Press; 1993.

11. Kelso JAS. \textit{Dynamic Patterns: The Self-Organization of Brain and Behavior}. Cambridge, MA: MIT Press; 1995.

12. Newell KM. Constraints on the development of coordination. In: Wade MG, Whiting HTA, editors. \textit{Motor Development in Children: Aspects of Coordination and Control}. Dordrecht: Martinus Nijhoff; 1986. p. 341--360.

13. Jolliffe IT, Cadima J. Principal component analysis: A review and recent developments. \textit{Philos Trans A Math Phys Eng Sci}. 2016;374(2065):20150202. doi:10.1098/rsta.2015.0202.

14. Bengio Y, Courville A, Vincent P. Representation learning: A review and new perspectives. \textit{IEEE Trans Pattern Anal Mach Intell}. 2013;35(8):1798--1828. doi:10.1109/TPAMI.2013.50.

15. Bommasani R, Hudson DA, Adeli E, Altman R, Arora S, von Arx S, et al. On the Opportunities and Risks of Foundation Models. arXiv. 2021. doi:10.48550/arXiv.2108.07258.

16. Ha D, Schmidhuber J. World Models. arXiv. 2018. doi:10.48550/arXiv.1803.10122.

17. LeCun Y. A Path Towards Autonomous Machine Intelligence. OpenReview. 2022.

18. Coravos A, Khozin S, Mandl KD. Developing and adopting safe and effective digital biomarkers to improve patient outcomes. \textit{NPJ Digit Med}. 2019;2:14. doi:10.1038/s41746-019-0090-4.

19. Topol EJ. High-performance medicine: the convergence of human and artificial intelligence. \textit{Nat Med}. 2019;25(1):44--56. doi:10.1038/s41591-018-0300-7.

20. Rajkomar A, Dean J, Kohane I. Machine learning in medicine. \textit{N Engl J Med}. 2019;380(14):1347--1358. doi:10.1056/NEJMra1814259.

21. Kelly CJ, Karthikesalingam A, Suleyman M, Corrado G, King D. Key challenges for delivering clinical impact with artificial intelligence. \textit{BMC Med}. 2019;17(1):195. doi:10.1186/s12916-019-1426-2.

22. Raynal J, Slangen P, Raynal E, Margerit J. Observable Performance Does Not Fully Reflect Adaptive System Organization: A Multi-Level Analysis of Gait Dynamics Under Occlusal Constraint. arXiv. 2026. doi:10.48550/arXiv.2605.00778.

23. Raynal J, Slangen P, Raynal E, Margerit J. From Static Representational Ambiguity to Observed Longitudinal Displacement: A Multi-Level Analysis of Gait Dynamics Under Occlusal Constraint. arXiv. 2026. doi:10.48550/arXiv.2605.13893.

24. Raynal J, Slangen P, Raynal E, Margerit J. From Observed Longitudinal Displacement to Internal Approximation: A Single-Subject Representation-Space Analysis of Gait Dynamics Under Occlusal Constraint. arXiv. 2026. doi:10.48550/arXiv.2605.15862.

25. Raynal J, Slangen P, Raynal E, Margerit J. From Performance to Representational Adequacy: A Representational Bootstrap Framework for Adaptive Biological Systems. arXiv. 2026. doi:10.48550/arXiv.2606.01374.

26. Kaprekar DR. An Interesting Property of the Number 6174. \textit{Scripta Mathematica}. 1955;21:244--245.

27. G\"odel K. \"Uber formal unentscheidbare S\"atze der Principia Mathematica und verwandter Systeme I. \textit{Monatshefte f\"ur Mathematik und Physik}. 1931;38:173--198. doi:10.1007/BF01700692.

28. Rosser JB. Extensions of Some Theorems of G\"odel and Church. \textit{Journal of Symbolic Logic}. 1936;1(3):87--91. doi:10.2307/2269028.

\end{document}